\documentclass[sigconf]{acmart}
\usepackage{multirow}
\AtBeginDocument{%
  \providecommand\BibTeX{{%
    \normalfont B\kern-0.5em{\scshape i\kern-0.25em b}\kern-0.8em\TeX}}}

\setcopyright{acmcopyright}
\copyrightyear{2022}
\acmYear{2022}
\acmDOI{XXXXXXX.XXXXXXX}

\acmPrice{15.00}
\acmISBN{978-1-4503-XXXX-X/18/06}

\begin{document}

\title{Effective Urban Region Representation Learning Using Heterogeneous Urban Graph Attention Network (HUGAT)}


\author{Namwoo Kim}
\email{ih736x@kaist.ac.kr}
\affiliation{%
  \institution{Korea Advanced Institute of Science and Technology}
  \streetaddress{P.O. Box 1212}
  \city{Daejeon}
  \country{South Korea}
}

\author{Yoonjin Yoon}
\authornote{Corresponding Author.}
\email{yoonjin@kaist.ac.kr}
\affiliation{%
  \institution{Korea Advanced Institute of Science and Technology}
  \streetaddress{P.O. Box 1212}
  \city{Daejeon}
  \country{South Korea}
}

\begin{abstract} 
Revealing the hidden patterns shaping the urban environment is essential to understand its dynamics and to make cities smarter. Recent studies have demonstrated that learning the representations of urban regions can be an effective strategy to uncover the intrinsic characteristics of urban areas. However, existing studies lack in incorporating diversity in urban data sources. In this work, we propose \textit{heterogeneous urban graph attention network} (HUGAT), which incorporates heterogeneity of diverse urban datasets. In HUGAT, \textit{heterogeneous urban graph} (HUG) incorporates both the geo-spatial and temporal people movement variations in a single graph structure. Given a HUG, a set of meta-paths are designed to capture the rich urban semantics as composite relations between nodes. Region embedding is carried out using \textit{heterogeneous graph attention network} (HAN). HUGAT is designed to consider multiple learning objectives of city's geo-spatial and mobility variations simultaneously. 
In our extensive experiments on NYC data, HUGAT outperformed all the state-of-the-art models. Moreover, it demonstrated a robust generalization capability across the various prediction tasks of crime, average personal income, and bike flow as well as the spatial clustering task.
\end{abstract}



\maketitle

\section{INTRODUCTION}

Learning effective urban region representation can provide valuable information that is crucial for many urban applications. Several region embedding models have been proposed to project the rich semantics of urban regions in the lower-dimensional latent vector space. Information regarding geo-spatial variations such as point-of-interest (POI) locations, land usage, and infrastructures has been indispensable in region embedding models. Encoder-decoder framework \cite{fu2019efficient, zhang2019unifying} and graph neural networks \cite{zhang2021multi, liu2020learning} were utilized to learn the region embeddings based on spatial relationship. There also has been notable efforts to incorporate temporal variations as well, using skip-gram objective \cite{wang2017region} and positive point-wise mutual information \cite{yao2018representing}. 

Although the aforementioned models demonstrated a level of efficiency in various tasks such as crime and economic growth predictions, and land usage classification, they still lack in capturing the rich semantics resulting from interactions among the diverse urban elements. For example, one leaves home at 7 a.m., picks up coffee at Starbucks at 7:10 a.m., and takes a taxi to the office in downtown to arrive at 8 a.m. The trip chain includes three different types of objects (region, POI, and time) and a series of human movement that connect them. One can model such data in separate spatial and temporal graphs \cite{wang2017region}, or simply consider human mobility as the co-occurrences between timestamp and physical regions in \cite{yao2018representing}. However, it is inevitable that deeper semantics are lost.

The key challenges is that dynamic urban environment naturally requires to consider multiple set of data source at the same time. The key data sources such as census tract, point-of-interest (POI) locations and category, and people movement records are available in silos and in their own taxonomy. Such diversity is not limited to data schema but also in their temporal dynamicity. For example, census tracts don't change everyday whereas the taxi trip data contains hundreds of thousands timestamps for daily pick-ups and drop-offs. 

In this paper, we propose a novel and efficient region embedding model called \textit{heterogeneous urban graph attention network} (HUGAT). In the model, we first define \textit{urban heterogeneous information network} (urban-HIN) built on \textit{heterogeneous urban graph} (HUG). HUG is a heterogeneous graph consisting of multiple types of urban nodes connected through diverse relations. A set of meta-paths defines the composite relations between nodes, and the meta-path based neighbors of a region identifies the set of regions that are connected through such semantics.

In HUG, we propose to use two categories of relations between node - the spatial and temporal relations. The spatial relation captures the spatial variations such as region adjacency and POI membership. The temporal relation captures interactions among regions realized through human movement such as taxi trips and POI check-ins. Given a HUG, meta-paths now can characterize the spatial, temporal and spatio-temporal semantics. In other words, urban-HIN is capable of incorporating both the spatial and temporal variations of urban regions simultaneously. 

Given a urban-HIN, the region embedding is carried out using \textit{heterogeneous graph attention network} (HAN) \cite{wang2019heterogeneous} based on a two-level (node-level and semantic-level) attentions by aggregating features from meta-path-based neighbors. Multi-task learning is designed to learn the deep semantics of the spatial and temporal patterns based on a set of region attribute distributions including trip, check-ins, and land usage. The goal is to project the regions of high similarity geometrically closer in the latent vector space, while dissimilar ones are far apart, by learning the regions' pairwise similarities and their interactions through heterogeneous information.

When the extensive experiments were carried out on New York City data, HUGAT outperformed all baseline models in diverse downstream tasks, including three prediction tasks (crime, average personal income, bike flow) as well as a region clustering task.
In addition, the results were evaluated for the influence of meta-paths, and qualitative assessment was carried out by inspecting the sets of closest regions in the latent vector space. 

In summary, our major contributions are highlighted as follows:

• We propose \textit{heterogeneous urban graph attention network} (HUGAT), a novel urban region representation learning model based on urban heterogeneous information networks (urban-HIN). To the best of our knowledge, it is the first model to exploit both heterogeneous graph and meta-path relations in urban region representation learning.

• The urban heterogeneous information network (urban-HIN) is able to incorporate both the geo-spatial and temporal composite relations between nodes to capture the rich urban semantics of spatial, temporal, and spatio-temporal nature. 

• The extensive experiments on NYC data demonstrate that HUGAT outperforms all the existing state-of-the-art baseline models in crime, average personal income, bike flow predictions, as well as the spatial clustering tasks by over 17\%, 18\%, 2.6\%, and 4.1\%, respectively. Such results indicate that HUGAT is capable to generalize across the diverse prediction and clustering tasks.

\section{RELATED WORK}
\subsection{Graph Embedding}
The graph embedding aims to learn a low-dimensional vector representation while preserving the network structure and property so that the embedding results can be used for diverse downstream tasks such as node classification, and node clustering. Algorithms such as DeepWalk \cite{perozzi2014deepwalk}, LINE \cite{tang2015line}, and Node2vec \cite{grover2016node2vec} generate random walks to capture local structures and then use skip-gram objective to learn node embeddings. However, these methods have limitations in capturing global topological structure as they rely solely on the local topology of the graph. Additionally, those methods cannot propagate feature information. 

Recently, graph neural networks (GNN) have shown 
promising representation capability. GNNs fall into two major categories - spectral-based and spatial-based Graph Neural Networks (GNNs). Spectral-based GNNs such as ChebNet \cite{defferrard2016convolutional} and Graph Convolution Network \cite{kipf2016semi} perform graph convolutions in the Fourier domain. Spatial-based GNNs such as Graph Isomorphism Network \cite{xu2018powerful}, GraphSAGE \cite{hamilton2017inductive}, and Graph Attention Network \cite{velivckovic2017graph} have been proposed to perform graph convolution operations directly in the graph domain. 

However, many real-world problems including urban region embedding are inherently heterogeneous, having more than one type of node or edge. Since all GNNs mentioned above are built for homogeneous graphs, they cannot be naturally adapted to heterogeneous graphs.

\subsection{Heterogeneous Network Embedding}
By utilizing the multiple types of edges and nodes, heterogeneous information network (HIN) \cite{sun2013mining} contains richer semantics and structural information compared to homogeneous network. There have been notable efforts to learn embedding of HIN. Metapath2vec \cite{dong2017metapath2vec} designed a single meta-path-based random walk and performed heterogeneous graph embedding using skip-gram objective. HERec \cite{shi2018heterogeneous} used meta-path-based neighbors to convert a heterogeneous graph into meta-path based homogeneous graphs, and performed node embedding using random walk. Instead of using random walk, Esim \cite{shang2016meta} generated the positive and negative meta-path instances to learn node embedding and importance of multiple meta-paths. HIN2Vec \cite{fu2017hin2vec} carried out multiple prediction training tasks to learn both the node embedding and meta-paths. 

Since each node pair can have multiple types of relationships, multiplex heterogeneous networks embedding such as MVE \cite{qu2017attention}, GATNE \cite{cen2019representation}, MNE \cite{zhang2018scalable} have been also proposed. MVE \cite{qu2017attention} used attention mechanism to learn view-specific node embedding. GATNE \cite{cen2019representation} performed node embedding using both base embedding and edge embedding simultaneously. MNE \cite{zhang2018scalable} learned node embedding by using a common embedding layer for several embedding of each relation edge type. HEER \cite{shi2018easing} attempted to bridge the semantic gap between heterogeneous nodes by learning additional edge representation of a heterogeneous information network. Recent surge in studies utilizing attention mechanism also includes heterogeneous information network embedding. HAN \cite{wang2019heterogeneous} designed a two-level (node-level and semantic-level) attentions to generate node embedding by aggregating features from meta-path-based neighbors. Based on HAN architecture, MAGNN \cite{fu2020magnn} considered intermediate node along the meta-paths. 

\subsection{Urban Region Embedding}
As the availability and scale of urban data sources various have expanded over time, research efforts on representations learning of urban regions have flourished in recent years. The key learning objective is to learn the spatial similarities so that regions of higher similarity to be geometrically closer in the latent space. 

Many studies explored region embedding for predictions of region specific features, including sociodemographic feature prediction \cite{wang2020urban2vec, jean2019tile2vec}, crime prediction \cite{wang2017region, zhang2021multi}, economic growth prediction \cite{hui2020predicting}, and land-usage classification \cite{yao2018representing, zhang2021multi}. There exist studies utilizing mobility data to model the association of regions based on people movement based on pointwise mutual information \cite{yao2018representing} or skip-gram objectives \cite{zhang2017regions}. Fu et al.  \cite{fu2019efficient} proposed a region embedding model that learns intra-region and inter-region similarities by using the intra-region POI network and the spatial autocorrelation layer. \cite{zhang2021multi} considers multi-view region similarities based on taxi trip records, POI check-in volumes, and the POI category distributions of regions. However, existing studies have limitations in integrating both spatial and temporal associations between regions.

\section{PRELIMINARY}
In this section, we give formal definitions of some important terminologies related to our work, and the problem statement. 

\textbf{Definition 3.1 Heterogeneous information network (HIN) \cite{sun2013mining}}
 A heterogeneous information network consists of three elements; heterogeneous graph, meta-paths, and meta-path neighbors. A heterogeneous graph is a graph \begin{math}
 \mathcal{G}=\ (\mathcal{V},\ \mathcal{E})
 \end{math} with multiple types of nodes and edges, and the node mapping functions \begin{math}\>\varphi:\mathcal{V}\rightarrow\mathcal{A}\end{math}  and the edge mapping function and \begin{math}\psi:\mathcal{E}\rightarrow\mathcal{R}\end{math} . \begin{math}\mathcal{A}\end{math} and \begin{math}\mathcal{R}\end{math} denote the sets of predefined node types and edge types, where \begin{math}\left|\mathcal{A}\right|+\left|\mathcal{R}\right|>2\end{math}. 

In HIN, two different nodes of a heterogeneous graph can be related through via composite node relations called \textit{meta-path}.

\textbf{Definition 3.2 Meta-path \cite{sun2011pathsim}} A meta-path \begin{math}\mathcal{M}\end{math} is denoted in the form of $A_1\xrightarrow{\text{$R_1$}}$
$A_2\xrightarrow{\text{$R_2$}}\ldots$ 
$A_{l-1}\xrightarrow{\text{$R_{l-1}$}}A_l$ (abbreviated as $A_1A_2...A_l$), wherein \begin{math}
R=R_1\circ R_2\circ\ldots\circ R_{l-1}
\end{math} describes the composite relations between node types $A_1$ and $A_l$.

Based on meta-path, one can identify the set of neighbors of each node called \textit{meat-path based neighbors}.

\textbf{Definition 3.3 Meta-path based neighbors \cite{wang2019heterogeneous}} A meta-path based neighbors $\mathcal{N}_i^\mathcal{M}$ of node $i$ and a meta-path $\mathcal{M}$ is the set of nodes directly connected with node i via meta-path $\mathcal{M}$.

\textbf{Definition 3.4 Urban region representation learning} Urban region representation learning aims to learn distributed and dense representations of each urban region $R={\{r_i\}_{i=1}^{N}}$, which is denoted as:
\begin{equation}
    Z=\{z_1,z_2,…,z_N\}, z_i\in \mathbb{R}^d
\end{equation}
Here, $N$ is the number of regions, $z_i$ is the embedding of the $r_i$, and every region is embedded into $d$-dimensional vector space.

\section{THE PROPOSED FRAMEWORK}

In this section, we present details of HUGAT. 

\begin{figure}
  \centering
  \includegraphics[width=0.9\columnwidth]{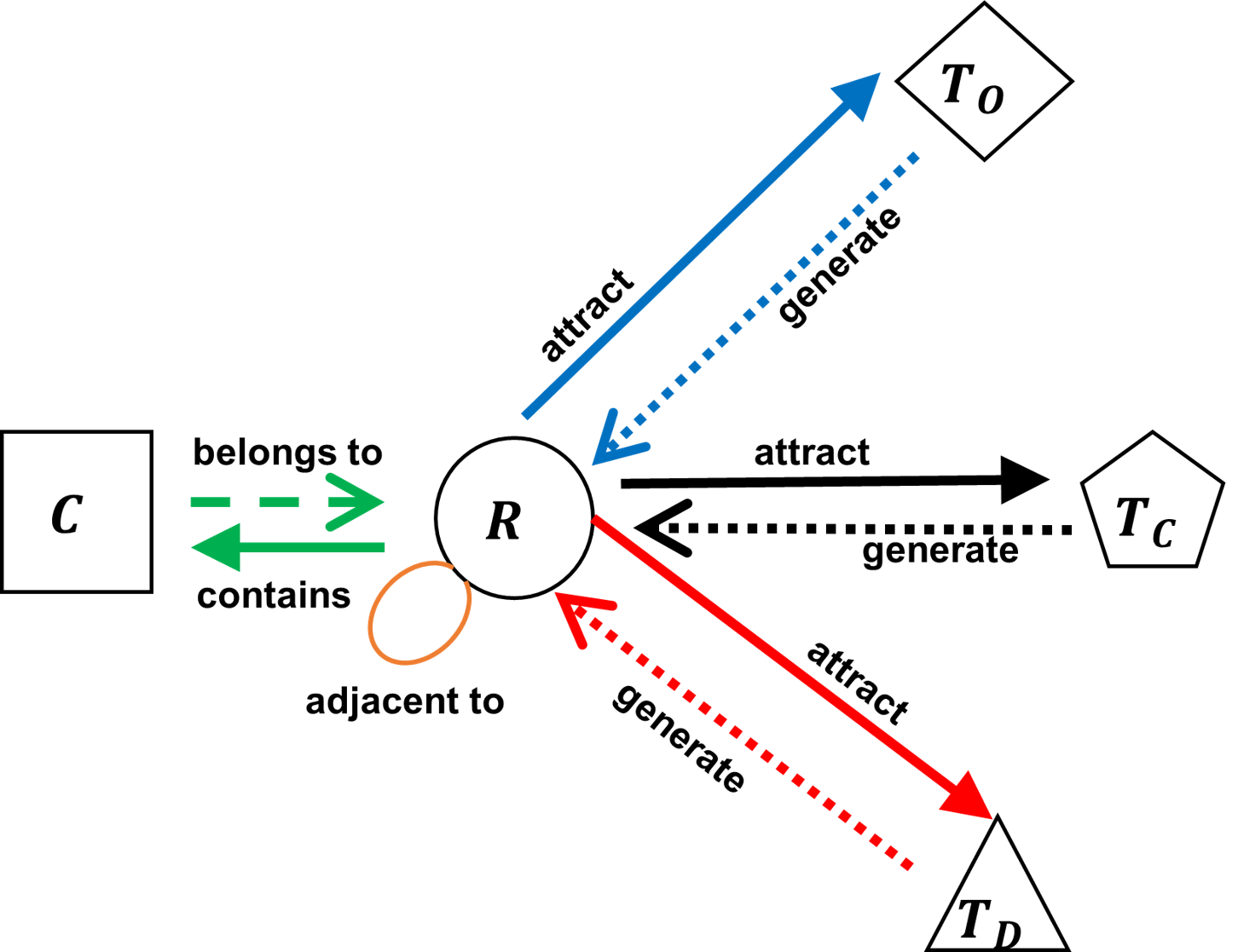}
  \caption{Graph schema of heterogeneous urban graph (HUG). There are five node types - \textbf{$R$} (region), \textbf{$C$} (POI category), $T_C$ (POI check-in time), $T_O$ (hot hours for taxi departure), $T_D$ (hot hours for taxi arrival). Edges represent two spatial relations (adjacentTo, contains/locatedIn), and temporal relations (attract/generate) of three time-dependent mobility events. }
  \Description{hug}
  \label{fig:scheme}
\end{figure}

\subsection{Urban Heterogeneous Information Network (urban-HIN)}
 Urban-HIN consists of three components – a heterogeneous urban graph (HUG), meta-paths, and meta-path-based neighbors. 

\textbf{Definition 4.1 Heterogeneous urban graph (HUG)} is defined as a directed graph  \begin{math}\mathcal{G}=\ (\mathcal{V},\ \mathcal{E}) \end{math}. Both the set of vertices \begin{math}\mathcal{V}\end{math} and the set of edges \begin{math}\mathcal{E}\end{math} may contain multiple spatial or temporal types. The node mapping function $\>\varphi:\mathcal{V}\rightarrow\mathcal{A}$ maps each node $v\in V$ to one of the node type $a\in A$. The edge mapping function and $\psi:\mathcal{E}\rightarrow\mathcal{R}$ maps the edge $e\in\mathcal{E}=\{(i,j)|i,j\in V\}$ to one of the relation type $r\in\mathcal{R}$. Since it is a directed graph, the reverse relations exist and \begin{math}\psi\left(\left(i,j\right)\right)\neq\psi\left(\left(j,i\right)\right)\end{math} in general.

In this study, HUG is constructed with five node types; region $R$, POI category $C$,check-in time interval $T_C$, trip origin time interval $T_O$, and trip destination time interval $T_D$. The edge types concerning static geospatial relations such as spatial adjacency (\textit{adjacentTo}) and POI membership (\textit{contains, locatedIn}) are straightforward to describe. However, the time-dependent relation types require a bit more elaboration. In the time-dependent data such as POI check-in, what relates the region and check-in time is the human movement. Such characteristics differs from the simple spatial adjacency in its complexity. To correctly model such property, we introduce the ‘\textit{generate}’ and ‘\textit{attract}’ relations between region and time. When a check-in occurred in region $R$ at time $T_C$, $R$ attracted a POI check-in at $T_C$ whereas $T_C$ generated a POI check-in at $R$. Likewise, $R$ attracted taxi pick-up events at time $T_O$ whereas $T_C$ generated the taxi pick-ups in $R$. Note that for taxi trips, only the hotspots with the top $k$\% origins and destinations are considered. In Figure ~\ref{fig:scheme}, the graph schema of HUG is illustrated with the vertex and edge types as described above.

\begin{figure*}
  \centering
  \includegraphics[width=\textwidth]{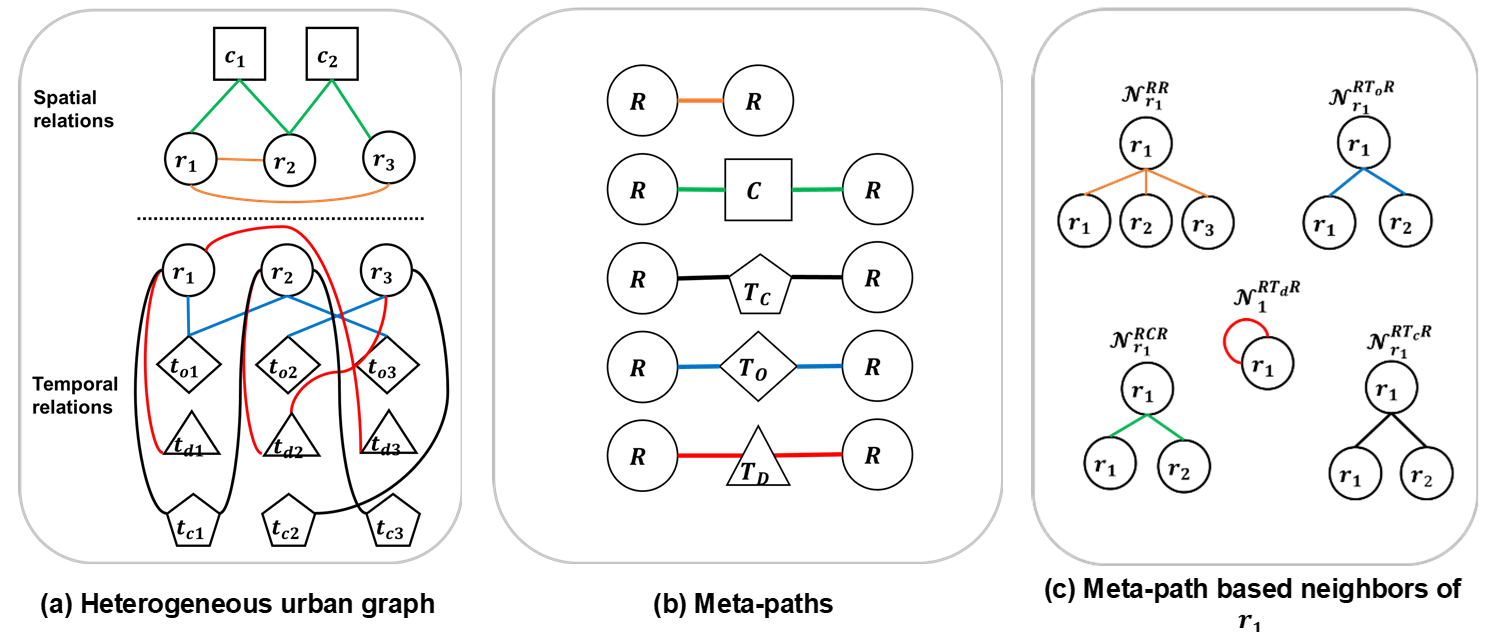}
  \caption{urban heterogeneous information network (urban-HIN). (a) heterogeneous urban graph (HUG) with five node types: $R$ (region), $C$ (POI category), $T_C$ (POI check-in time), $T_O$ (hot hours for taxi departure), $T_D$ (hot hours for taxi arrival)), (b) meta-paths, (c) The meta-paths based neighbors of $r_1$.}
  \Description{ha to dense}
  \label{fig:hug}
\end{figure*}

\textbf{Definition 4.2 urban heterogeneous information network (urban-HIN)}
 The urban heterogeneous information network (urban-HIN) is a HIN constructed upon heterogeneous urban graph (HUG). An example of urban heterogeneous information network (urban-HIN) is shown in Figure ~\ref{fig:hug}. Given a heterogeneous urban graph (Figure ~\ref{fig:hug} (a)), five meta-paths define semantic relations among regions, including, $RR$, $RCR$, $RT_OR$, $RT_DR$, and $RT_CR$ (Figure ~\ref{fig:hug} (b)). The meta-path based neighbors of $r_1$ identify the regions connected through the meta-paths  (Figure ~\ref{fig:hug} (c)). One can observe certain disjoint regions now belong to the same meta-path based neighbor set. For instance, $r_2$ and $r_3$ are spatially disjoint but become neighbors by having common POI category $c_2$ based on the meta-path $RCR$. Similarly, $r_2$ and $r_3$ become neighbors based on $RT_DR$ representing two destination hot-regions in a given time. 

\subsection{Regional Attribute Distribution and Inter-Relationship}
In this section, we define regional attribute distributions including trip distribution, check-in distribution, and land usage distribution. 
\\
\textbf{Trip distribution } Based on the real-world trip dataset, we can construct the origin-destination matrix $F$. Then, the conditional distribution of origin region $r_i$ for a given destination region $r_j$ is defined as:
\begin{equation}
  p_{org|dst}\left(org = r_i\middle|dst = r_j\right)=\frac{F_{r_i,r_j}}{F_{:,r_j}}
\end{equation}
, where $F_{r_i,r_j}$ is the trip frequency from $r_i$ to $r_j$, and $F_{:,r_j}$ is the total number of inbound trips to $r_j$.  $p_{dst|org}$ can be obtained in the similar manner.
$p_{dst|org}\in\mathbb{R}^{N\times N}$ and $p_{org|dst}\in\mathbb{R}^{N\times N}$ represents the pairwise regional associations through the semantics of trip origin and destination.
\\

\textbf{Check-in distribution } The probability distribution of check-in categories of region $r_i$ is defined as: 
\begin{equation}
 p_{cat|reg}\left(cat = c_k\middle|reg = r_i\right) = \frac{V_{c_k,r_i}}{V_{:,r_i}}
\end{equation}
, where $V_{c_k,r_i}$ is the number of visits to a POI of type $c_k$ in $r_i$. To model the relationship between regions based on POI check-in activity, we define the similarity matrix \begin{math}\mathcal{S}_{chk}\in\mathbb{R}^{N\times N}\end{math}. The similarity of check-in activities between regions $r_i$ and $r_j$ is obtained using Hellinger distance \cite{1089532} of the two discrete probability distributions $p_{cat|reg}(:,r_i)$ and  $p_{cat|reg}(:,r_j)$ as:
\begin{equation}
\begin{split}
    \mathcal{S}_{chk}\left(r_i,r_j\right)  &=\frac{1}{\sqrt2}\parallel\sqrt{p_{cat|reg}(:,r_i)}-{\sqrt{p_{cat|reg}(:,r_j)}\parallel}_2 \\
    & = \frac{1}{\sqrt{2}}\sqrt{\sum_{c_{k}\in C}(\sqrt{p_{cat|reg}(c_k,r_i)}-\sqrt{p_{cat|reg}(c_k,r_i)})^2}
\end{split}
\end{equation}
\\
\textbf{Land usage distribution } Land usage distribution of region $r_i$ is defined as:
\begin{equation}
    p_{type|reg}\left(type=l_m\middle|reg=r_i\right)=\frac{A_{l_m,r_i}}{A_{:,r_i}}
\end{equation}
where, $A_{l_m,r_i}$ is the area occupied by the land use type $l_m$ in $r_i$. Then, similarity matrix for land use \begin{math}\mathcal{S}_{land}\in\mathbb{R}^{N\times N}\end{math} is defined in a similar manner in  eq.4. 
\subsection{Heterogeneous Graph Attention Network}
Once the urban-HIN is constructed, HAN \cite{wang2019heterogeneous} was used for region representation learning. For simplicity, we denote region $r_i$ as $i$ and its $m$-dimensional feature vector $h_i \in \mathbb{R}^m$.  Since HUG consists of multiple node types with different feature spaces, the node type specific projection matrix \begin{math}\mathcal{M_\varphi}\end{math} is applied to map them into the same feature space as follows:

\begin{equation}
    h_i^\prime=\mathcal{M_\varphi} h_i
\end{equation}

Then, attention mechanism is applied to learn the weights of the meta-path based neighbors $\mathcal{N}_i^{\mathcal{M}_k}$. The attention score between two nodes connected by the meta-path is defined as:

\begin{equation}
    \alpha_{ij}^{\mathcal{M}_k}=\frac{exp\left(f_{\mathcal{M}_k}\left(h_i^\prime,h_j^\prime\right)\right)}{\sum_{l\in\mathcal{N}_i^{\mathcal{M}_k}} e x p\left(f_{\mathcal{M}_k}\left(h_i^\prime,h_l^\prime\right)\right)},\forall l\in\mathcal{N}_i^{\mathcal{M}_k}
\end{equation}
, where $f_{\mathcal{M}_k}(\cdot,\cdot)$ is the neural network that performs the node-level attention for the given meta-path $\mathcal{M}_k$. $\alpha_{ij}^{\mathcal{M}_k}$ is the importance of node $i$ to the node $j$, and $\mathcal{N}_i^{\mathcal{M}_k}$ is the meta-path based neighbors of node $i$ given ${\mathcal{M}_k}$.
Aggregation process of node $i$ is followed by aggregating over the meta-path based neighbor’s features with the corresponding attention score: 

\begin{equation}
    y_i^{\mathcal{M}_k}=\Vert_{l=1}^ K\sigma\left(\sum_{j\in\mathcal{N}_i^\mathcal{M}}{\alpha_{ij}^{\mathcal{M}_k} h_j^\prime}\right)
\end{equation}
, where $\sigma(\cdot)$ is a nonlinear function, $y_i^{\mathcal{M}_k}$ is the embedding of the node $i$ for the given meta-path $\mathcal{M}_k$, and $K$ is the number of heads.

Given the meta-path set \{$\mathcal{M}_1,\mathcal{M}_2,\ldots,\mathcal{M}_P$\}, $P$ groups of node embeddings of corresponding meta-paths  \{${\mathbf{Y}}^{\mathcal{M}_1},\mathbf{Y}^{\mathcal{M}_2},\ldots,\mathbf{Y}^{\mathcal{M}_P}$\} are obtained from node level attention. To learn the importance of different meta-paths, the importance of each meta-path $\mathcal{M}_k$ is calculated as:
\begin{equation}
    w_{\mathcal{M}_k}=\frac{1}{\left|\mathcal{V}\right|}\sum_{j\in\mathcal{V}}{q^Ttanh\left({W}y_j^{\mathcal{M}_k}+b\right)}
\end{equation}
where $q$ is a single layer feed forward network, $W$ is the weight matrix, $b$ is the bias. Then, $\beta^{\mathcal{M}_k}$ which denotes the weight of meta-path $\mathcal{M}_k$ can be obtained by normalizing $w_\mathcal{M}$ with softmax function.
\begin{equation}
\beta^{\mathcal{M}_k}=\frac{exp\left(w_{\mathcal{M}_k}\right)}{\sum_{j=1}^{P}exp\left(w_{\mathcal{M}_j}\right)}
\end{equation}

Now, we can fuse the node-level embeddings and the weights of the meta-path:
\begin{equation}
    \mathbf{Y}=\sum_{k=1}^{M}\beta^{\mathcal{M}_k}\mathbf{Y}^{\mathcal{M}_k}
\end{equation}
Finally, through a dense layer $g$, we obtain a final representation $Z=g(Y)$ by projecting the node embeddings into the vector space with the desired output dimensions. 

\begin{figure*}[!h]
  \centering
  \includegraphics[width=0.9\textwidth]{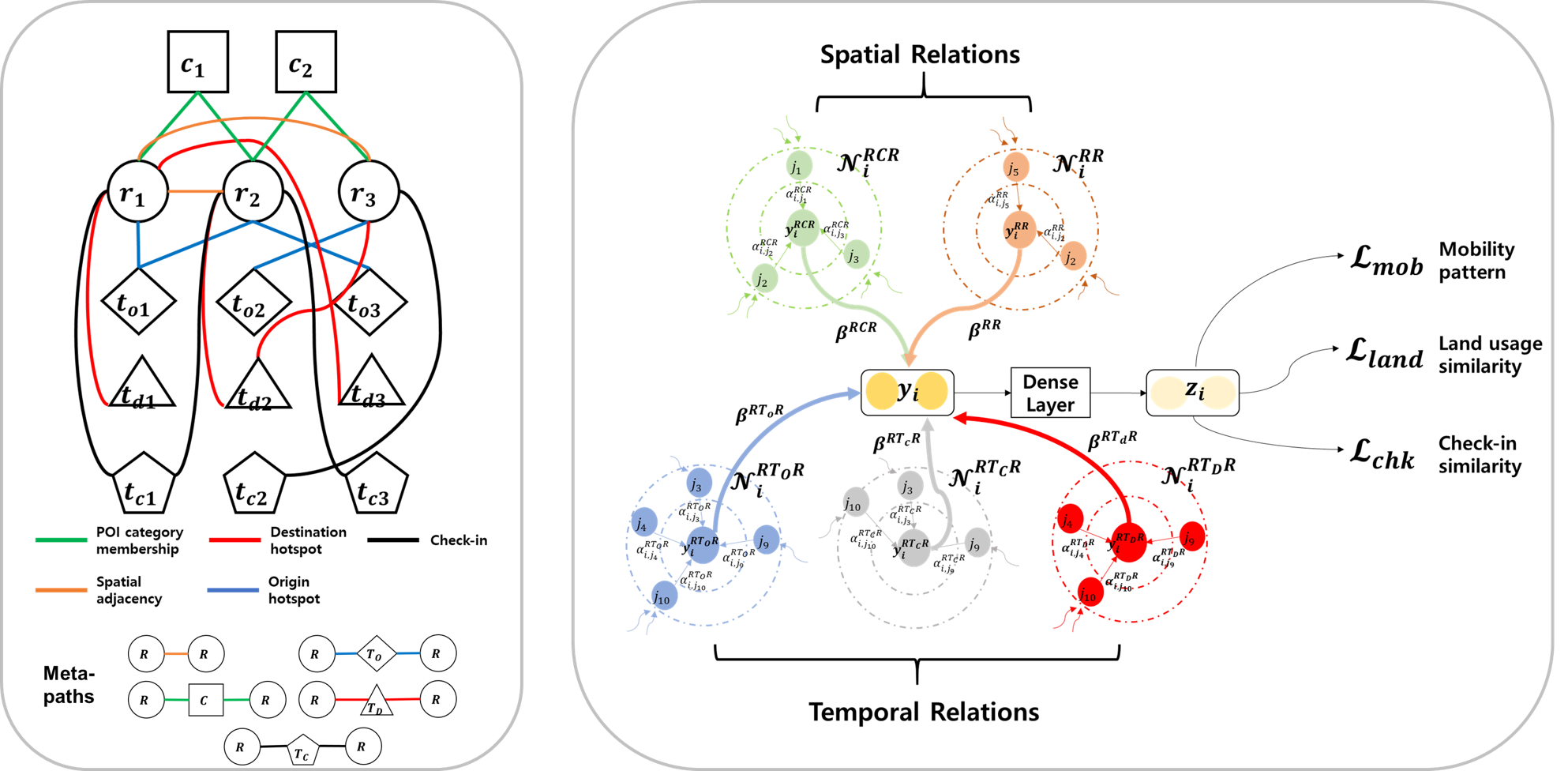}
  \caption{HUGAT Framework. Given the urban-HIN with 5 meta-paths \{\textbf{$RR,\ RCR,\ RT_OR,\ RT_DR,\ RT_CR$}\}, five corresponding groups of region embeddings \{\textbf{$\mathbf{Y}^{RR},\mathbf{Y}^{RCR},\mathbf{Y}^{RT_OR}, \mathbf{Y}^{RT_DR}, \mathbf{Y}^{RT_CR}$}\} are obtained. Semantic attention is applied to learn the importance of each meta-path. Final region representation is then obtained by using a dense layer $g$, projecting region embeddings into the desired output dimensions. In this study, HUGAT was designed to learn the pairwise similarities and interactions of regions based on the trip, check-in, and land usage distribution.} 
  \Description{ha to dense}
  \label{fig:framework}
\end{figure*}

\subsection{Learning Objectives}
We consider the following learning objectives to ensure that learned region embeddings retain regional similarities.
\\
\textbf{Mobility loss } Having the embeddings $Z$=\{$z_i$\}$_{i=1}^n$, we can model the estimated conditional distribution of origin regions for a given destination region $r_j$ as:
\begin{equation}
    {\hat{p}}_{org|dst}\left(r_i\middle| r_j\right)=\frac{exp\left({z_i}^Tz_j\right)}{\sum_{k} e x p\left({z_k}^Tz_j\right)\ \mathrm{\ } }
\end{equation} Likewise, ${\hat{p}}_{dst|org}$ is obtained. Then, we use the Kullback-Leibler divergence \cite{Kullback1951OnIA} to approximate well the estimated conditional trip distribution as:
\begin{equation}
    \begin{aligned}
       \mathcal{L}_{mob}=\sum_{\left(r_i,r_j\right)} K L\left(p_{org|dst}\left(r_i\middle| r_j\right),{\hat{p}}_{org|dst}\left(r_i\middle| r_j\right)\right)+\\
    \sum_{\left(r_i,r_j\right)} K L\left(p_{dst|org}\left(r_j\middle| r_i\right),{\hat{p}}_{dst|org}\left(r_j\middle| r_i\right)\right) 
    \end{aligned}
\end{equation}
Through $\mathcal{L}_{mob}$, our model can capture the mobility patterns.
\\
\textbf{Check-in loss} To learn the pairwise similarities among regions regarding POI check-ins and land usage, we define two loss terms of $\mathcal{L}_{chk}$ and $\mathcal{L}_{land}$. 
Given the embeddings $Z$=\{$z_i$\}$_{i=1}^n$, we can estimate the similarity of check-in activities between $r_i$\ and $r_j$ as:\ \begin{equation}
    {\hat{\mathcal{S}}}_{chk}\left(r_i,r_j\right)=\frac{1}{\sqrt2}\parallel\sqrt{\hat{p}_{cat|reg}(:,r_i)}-{\sqrt{\hat{p}_{cat|reg}(:,r_j)}\parallel}_2 
\end{equation}
where, $\hat{p}_{cat|reg}(:,r_i)$=softmax($z_i$). Then, the loss function can be formalized as:
\begin{equation}
    \mathcal{L}_{chk}={\sum_{\left(r_i,r_j\right)}\left|\mathcal{S}_{chk}\left(r_i,r_j\right)-{\hat{\mathcal{S}}}_{chk}\left(r_i,r_j\right)\right|}^2
\end{equation}
\textbf{Land usage loss}
Similarly, the land-use loss $\mathcal{L}_{land}$ is defined as: 
 \begin{equation}
    {\hat{\mathcal{S}}}_{land}\left(r_i,r_j\right)=\frac{1}{\sqrt2}\parallel\sqrt{\hat{p}_{type|reg}(:,r_i)}-{\sqrt{\hat{p}_{type|reg}(:,r_j)}\parallel}_2 
\end{equation}
\begin{equation}
    \mathcal{L}_{land}={\sum_{\left(r_i,r_j\right)}\left|\mathcal{S}_{land}\left(r_i,r_j\right)-{\hat{\mathcal{S}}}_{land}\left(r_i,r_j\right)\right|}^2
\end{equation}

Finally, the model learning objective function is defined as a convex combination of three loss terms $\mathcal{L}_{mob}$, $\mathcal{L}_{chk}$ and $\mathcal{L}_{land}$: 
\begin{equation}
    \mathcal{L}=\alpha\ast\mathcal{L}_{chk}+\beta\ast\mathcal{L}_{land}+\gamma\ast\mathcal{L}_{mob}
\end{equation}
In model training, we set $\alpha$ = 0.3, $\beta$=0.6, and $\gamma$=0.1.

For better understanding, the overall process of HUGAT, including two-level attention, final node representation, and learning objectives is shown in Figure ~\ref{fig:framework}.

\section{EXPERIMENTS}
In this section, we present the results after conducting extensive experiments on real-world datasets from New York City. 
\subsection{Data Description}
The geo-spatial unit is the census tract, and relevant datasets were collected from multiple sources as described below. 
\\
\textbf{Census tracts}\footnote{https://www.census.gov} A total of 282 census tract units were collected from 2010 decennial U.S. Census.
\\
\textbf{Taxi data}\footnote{https://www1.nyc.gov/site/tlc/index.page} Over 1.3 hundred million Yellow Cab taxi trip records in Manhattan, New York in year 2012 were used. Each trip record contains pick-up/drop-off timestamp, locations, and trip distance.
\\
\textbf{Check-in \& POIs data} The Foursquare check-in data \cite{10.1145/2814575} was used to track the POI category and check-in times. The dataset includes check-in records in NYC for 17 months between Apr. 2012 and Sep. 2013. Each check-in record contains anonymized user ID, timestamp, venue ID, and its category. A total of 334,456 check-in records from 14,723 users on 30,587 venues across 9 root categories in Manhattan, NY were utilized.
\\
\textbf{PLUTO data}\footnote{https://www1.nyc.gov/site/planning/index.page} NYC Primary Land Use Tax Lot Output (PLUTO) data provides land use and infrastructure information. A total of 11 land use types were considered at the census tract level.
\\
\textbf{Crime data}\footnote{https://opendata.cityofnewyork.us/} Over one hundred thousand crime events including valid felony, violation and misdemeanor reported to NYPD in 2013 were extracted from NYPD Complaint data published in NYC OpenData. 
\\
\textbf{Income data}\footnote{https://www.census.gov/programs-surveys/acs} the average personal income data from the American Community Survey 5-Year Data 2013 was used. 
\\
\textbf{Bike data}\footnote{https://ride.citibikenyc.com/system-data} Monthly trips on Citi Bike, the NYC bike share system, were sourced between Jul. 2013 and Dec. 2013. Each trip record contains the time and the location of bike pick-up an drop-off.
\\
The summary statistics of New York City HUG constructed for our experiment is provided in Table \ref{tab:uhin}. In the New York City HUG, a 250 dimensional random vector was used as a node feature.

\begin{table}
  \caption{Statistics of NYC heterogeneous urban graph (HUG)}
  \label{tab:uhin}
\begin{tabular}{|c|c|c|c|c|} 
\hline
Relations (A-B) & \# A & \# B & \# A-B & Features              \\ 
\hline
$R-T\textsubscript{C}$            & 282  & 168  & 33041  &  \\ 
\cline{1-4}
$R-T\textsubscript{O}$            & 282  & 168  & 34149  &                       \\ 
\cline{1-4}
$R-T\textsubscript{D}$            & 282  & 168  & 38109  &     {250}                   \\ 
\cline{1-4}
$R-R$             & 282  & 282  & 1744   &                       \\ 
\cline{1-4}
$R-C$             & 282  & 9    & 2186   &                       \\
\hline
\end{tabular}
\end{table}

\subsection{Baselines}
Eight baseline models were used for performance comparison including two variants of the proposed model. Short descriptions of each model is provided below. 
\\
\textbf{Topology based graph embedding methods}
\begin{itemize}
\item{\textbf{DeepWalk}} \cite{perozzi2014deepwalk} learns node embeddings by modelling random walks. In the experiments, we generated walks on the spatial network, and set the length of walks = 90, embedding size = 32, and window size = 5.
\end{itemize}

\begin{itemize}
	\item {\textbf{Node2vec}} \cite{grover2016node2vec}  is an unsupervised graph learning model which learns node embeddings from graph structured data. We applied node2vec to learn the embeddings for each region on the spatial network. The model parameters were set as follows; length of walks = 90, embedding size = 32, window size =5, return parameter (p) = 0.25, in-out parameter (q) = 1. 
\end{itemize}

\textbf{State-of-the-art methods}
\begin{itemize}
	\item {\textbf{DGE}} \cite{wang2017region} is a region embedding approach that jointly learns node embedding incorporating mobility flow network and spatial network. The spatial network is weighted by inverse distance and mobility flow network is weighted by the number of flows. Then, DGE learns node embeddings through random walks generated from combined network via skip-gram objectives.
\end{itemize}

\begin{itemize}
	\item {\textbf{MVPNE}} \cite{fu2019efficient} uses intra-region POI networks, autocorrelation layer, and autoencoder framework to represent regions considering both intra-region and inter-region similarities. The size of embeddings was set as 32.
\end{itemize}

\begin{itemize}
	\item {\textbf{GMEL}} \cite{liu2020learning} captures spatial correlation using land usage, structural information, and commute flow data. In GMEL, two separate GAT layers were used for origin and destination of commute flow. We set the embedding size of each layer to 128 as suggested by the authors. 
\end{itemize}

\begin{itemize}
	\item {\textbf{MVURE}} \cite{zhang2021multi} jointly learns region embeddings based on graph attention network. MVURE models multi-view region correlations by utilizing taxi-trip records and inherent region attribute. We set the embedding size to 96 as suggested by the authors.
\end{itemize}

\textbf{Variants of our method}
\begin{itemize}
	\item {\textbf{HUGAT\textsubscript{mob}}} is a variant of our model considering only the mobility loss  $\mathcal{L}_{mob}$.
\end{itemize}

\begin{itemize}
	\item {\textbf{HUGAT\textsubscript{hom}}} is a variant of our model considering homogeneous spatial networks and GAT to generate embeddings.
\end{itemize}

\begin{table*}
\centering
\caption{Results on downstream applications}
\label{tab:prediction}
\begin{tabular}{|c|c|c|c|c|c|c|c|c|c|c|c|} 
\hline
\multirow{2}{*}{\begin{tabular}[c]{@{}c@{}}Model\end{tabular}} & \multicolumn{3}{c|}{\begin{tabular}[c]{@{}c@{}}Crime\\{[}counts]\end{tabular}} & \multicolumn{3}{c|}{\begin{tabular}[c]{@{}c@{}}Personal Income\\{[}\$1k]\end{tabular}} & \multicolumn{3}{c|}{\begin{tabular}[c]{@{}c@{}}Bike Flow\\{[}counts]\end{tabular}} & \multicolumn{2}{c|}{Spatial Clustering}  \\ 
\cline{2-12}
                                                                    & MAE              & RMSE             & R\textsuperscript{2}                     & MAE            & RMSE            & R\textsuperscript{2}                                & MAE             & RMSE            & R\textsuperscript{2}                           & NMI            & ARI                     \\ 
\hline
DeepWalk                                                            & 245.659          & 326.955          & -0.149                                   & 46.781         & 58.592          & -0.661                                              & 42.923          & 67.409          & 0.151                                          & 0.709          & 0.514                   \\ 
\hline
Node2Vec                                                            & 242.893          & 332.124          & -0.185                                   & 57.847         & 67.748          & -1.22                                               & 44.364          & 69.078          & 0.109                                          & 0.690          & 0.478                   \\ 
\hline
DGE                                                                 & 215.015          & 306.587          & -0.011                                   & 43.226         & 50.982          & -0.257                                              & 36.754          & 64.682          & 0.219                                          & 0.049          & 0.010                   \\ 
\hline
MVPNE                                                               & 208.119          & 300.491          & 0.03                                     & 36.845         & 45.499          & -0.001                                              & 38.087          & 66.037          & 0.186                                          & 0.159          & 0.029                   \\ 
\hline
GMEL                                                                & 225.538          & 315.783          & -0.073                                   & 22.06          & 30.298          & 0.556                                               & 47.2            & 71.403          & 0.048                                          & 0.531          & 0.296                   \\ 
\hline
MVURE                                                               & 204.108          & 283.63           & 0.134                                    & 22.028         & 29.739          & 0.572                                               & 37.517          & 62.011          & 0.282                                          & 0.668          & 0.458                   \\ 
\hline
HUGAT\textsubscript{hom}                                                            & 186.079          & 271.562          & 0.207                                    & 20.115         & 27.097          & 0.645                                               & 36.108          & 62.139          & 0.279                                          & 0.738          & 0.552                   \\ 
\hline
HUGAT\textsubscript{mob}                                                            & 167.585          & 250.081          & 0.328                                    & 18.03          & 24.699          & 0.705                                               & 36.097          & 61.801          & 0.287                                          & 0.74           & 0.558                   \\ 
\hline
HUGAT                                                               & \textbf{167.338} & \textbf{249.595} & \textbf{0.33}                            & \textbf{18.02} & \textbf{24.682} & \textbf{0.705}                                      & \textbf{35.751} & \textbf{61.347} & \textbf{0.297}                                 & \textbf{0.745} & \textbf{0.571}          \\
\hline
\end{tabular}
\end{table*}
\subsection{Experiment Setup}
Models including all baselines and HUGAT were implemented using PyTorch and Deep Graph Library \cite{wang2019dgl}. Our model employed one HAN layer with 10 heads was used with 128-dimensional hidden node features, and an additional dense layer to project the hidden features into 32-dimensions. The training epochs were set to 1000, and learning rate was set to 0.001. Adam optimizer was used to optimize the model. The result are presented as the average values of five random runs. 
\subsection{Results}
In this section, we present the results of four downstream tasks: crime prediction, personal income prediction, bike flow prediction, and region clustering. 
\subsubsection{prediction task}
In crime and average personal income predictions tasks, we used the Lasso regression model with 5-fold cross-validation  by using the learned region embeddings as features in the regression model. 
In the bike flow prediction, we trained the regression model ${\hat{f}}_{ij}$=$L(z_i,z_j,d_{ij})$, where, ${\hat{f}}_{ij}$ is the estimated flow between $r_i$ and $r_j$, and $d_{ij}$ is travel distance between region $r_i$ and $r_j$. The travel distance between the centroids of regions is measured using Open Source Routing Machine (OSRM) \cite{luxen-vetter-2011}. 

Prediction errors were measured in Mean Absolute Error (MAE), Root Mean Square Error (RMSE), and $R^2$ as presented in Table ~\ref{tab:prediction}. The results show that HUGAT not only outperforms all the other baselines, but also is capable to generalize across multiple prediction tasks. 

It is also notable that two variants of HUGAT also outperform the state-of-the-art models. One can infer that learning the pairwise similarities is critical to achieve such results in all variants HUGAT. Improvement of HUGAT\textsubscript{mob} over HUGAT\textsubscript{hom} clearly indicates the contribution of human mobility data in learning the effective region embedding.
Finally, the results that HUGAT outperforms HUGAT\textsubscript{hom} and HUGAT\textsubscript{mob} verifies  the merit of utilizing the deep semantics of diverse urban elements through urban-HIN, as well as the effectiveness of learning the pairwise similarities in the region embedding.

\subsubsection{clustering task}
In region clustering task, we applied k-means clustering \cite{Macqueen67somemethods} on the embeddings. We use the district division by the Community Boards as the ground truth, which divides the New York city into 12 districts. The clustering results were evaluated using two measures: Normalized Mutual Information (NMI) \cite{4749258} and Adjusted Rand Index (ARI) \cite{Hubert1985ComparingP}. Since the performance of k-means is affected by initialization of the centroids, we repeated the experiment 10 times, and report the average values in Table ~\ref{tab:prediction}. 

\begin{figure}
  \centering
  \includegraphics[width=\columnwidth]{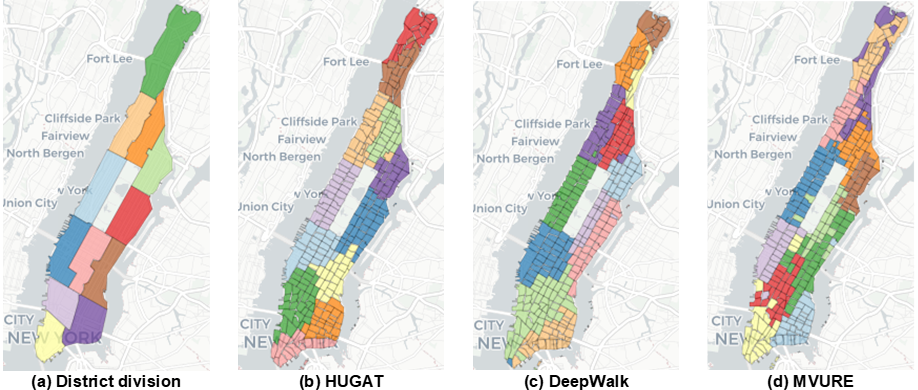}
  \caption{Visualization of clustering results.}
  \Description{Visualization of clustering results.}
  \label{fig:clustering}
\end{figure}

As shown in Table ~\ref{tab:prediction}, our method outperforms all baselines. Figure ~\ref{fig:clustering} also show that our results effectively delineate the boundaries of district division, indicating that our method effectively captures the structural information of spatial networks.

Another interesting observation is that topology based graph embedding models show better performance compared to the state-of-the-art baselines in clustering task. Since DeepWalk and Node2vec are designed to learn local topology of given spatial network, the spatial contiguity is well preserved in these models. 

\subsection{Impact of Meta-paths}
\begin{figure}
  \centering
  \includegraphics[width=\columnwidth]{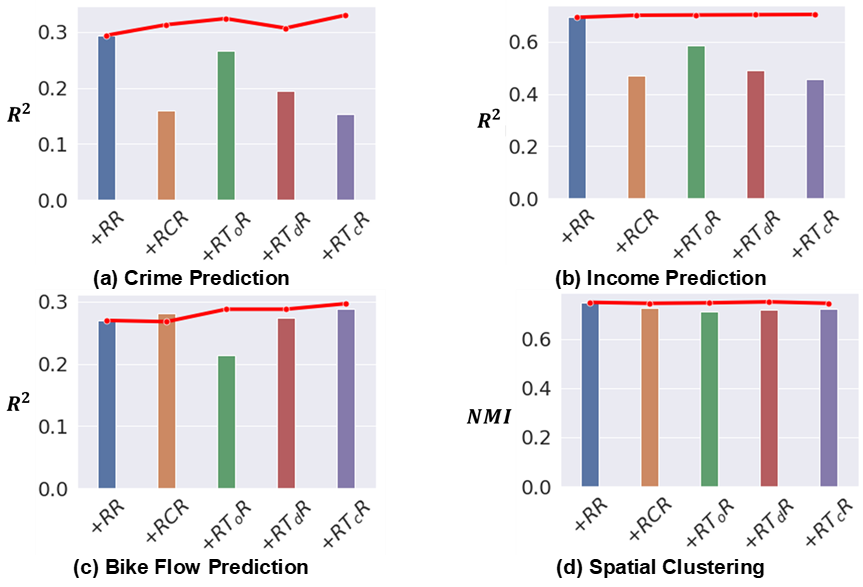}
  \caption{Evaluation of meta-paths. Single meta-path performance is shown in bar graph, and line graph shows the performance changes by incorporating the corresponding meta-path incrementally.}
  \Description{R2 & NMI}
  \label{fig:impact}
\end{figure}

In this section, we discuss the impact and validity of five meta-paths used in our model. Meta-paths are evaluated both individually and in combination by incorporating meta-paths one by one. In Figure ~\ref{fig:impact}, performance of individual meta-path is shown in bar graphs, and their incremental impact is shown in line graphs. ${R^2}$ of three prediction tasks and $NMI$ of the clustering task are shown. 

It is clear that there is no single best meta-path producing the best performance across the four downstream tasks. In addition, the performance of three prediction tasks were improved when all five meta-paths are considered. 

\subsection{Qualitative Assessment}
In this section, we discuss the qualitative aspects of embedding results by inspecting the nearest neighbors of region $r_3$ and $r_{151}$ in the latent space. 

\begin{figure}
  \centering
  \includegraphics[width=\columnwidth]{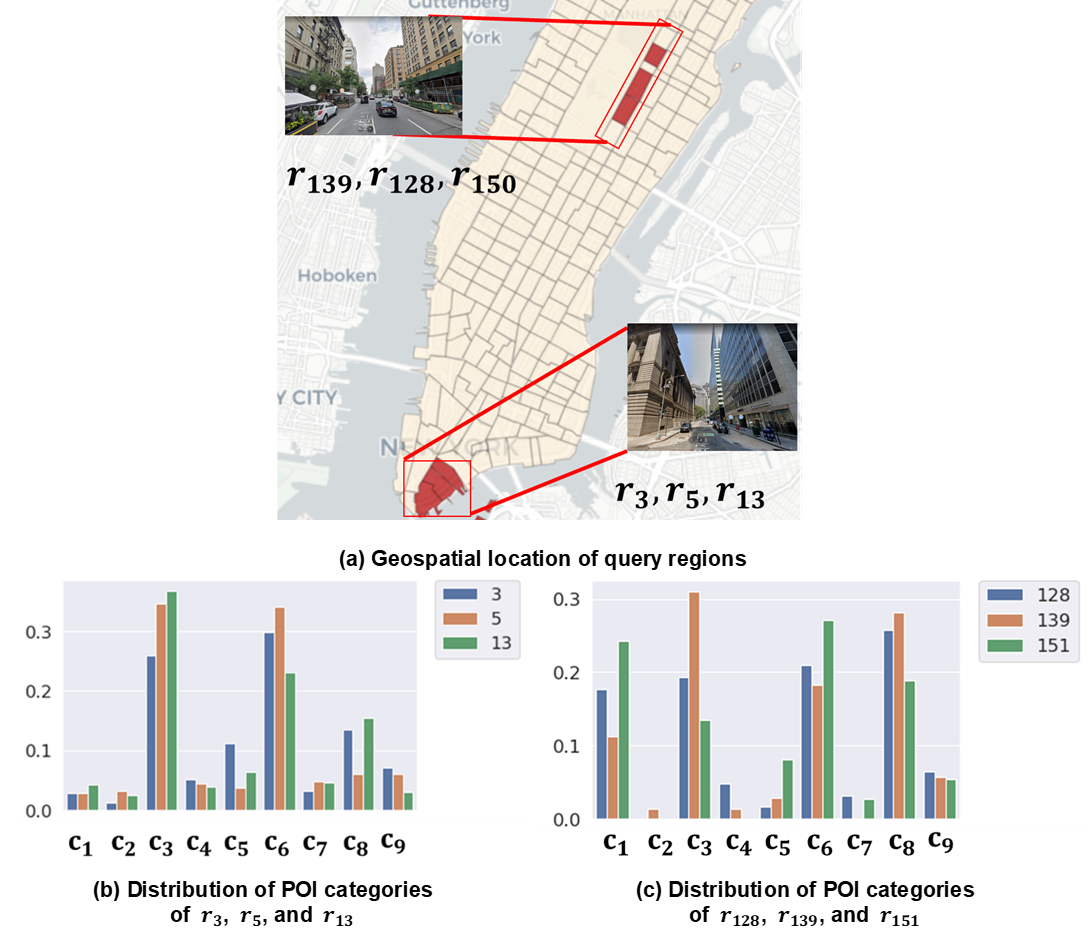}
  \caption{POI category distributions of the set of nearest neighbors in the latent space. (b-c) distribution of POI categories for query regions. $c_1$: Arts \& Entertainment, $c_2$: College \& University, $c_3$: Food, $C_4$: Nightlife Spot, $c_5$: Outdoors \& Recreation, $c_6$: Professional \& Other Places, $c_7$: Residence, $c_8$: Shop \& Service, $c_9$: Travel \& Transport.}
  \Description{poi distribution}
  \label{fig:qa}
\end{figure}

As shown in Figure \ref{fig:qa} (a), $r_3$ and its two nearest neighbors, $r_5$ and $r_{13}$, are located in Financial District in NYC. One can observe that POI category $c_3$ (Food) and $c_6$ (Professional \& Other places) are two main categories (Figure ~\ref{fig:qa} (b). One can also observe that the POI distributions of three regions are similar. One can make similar observation in $r_{151}$ located in the Upper East Side, when compared with its two nearest neighbors in the latent space $r_{128}$ and $r_{139}$.  

\section{CONCLUSION}
In this paper, we proposed \textit{heterogeneous urban graph attention network} (HUGAT) and the urban heterogeneous information network (urban-HIN). urban-HIN effectively captures the multiple urban elements and their diverse relations, as the deep semantics of urban space and human movement are captured through meta-paths . Heterogeneous attention network (HAN) provided an effective mean to learn the region embedding both at the node and the semantic level. By learning the pairwise region similarities and interactions, the model effectively learns the deep semantics and their interactions between regions. 
When New York City dataset is applied, our model outperforms all baseline models in all downstream tasks, including prediction tasks of crime, average personal income and bike flow, as well as the spatial clustering task. The model's experiment results and its robust generalization capability provide a strong set of evidence that utilizing diverse data sources is critical, and HUGAT is an effective framework to achieve the purpose.

\begin{acks}
This work was supported by the National Research Foundation of Korea (NRF) Basic Research Program grant funded by the Korea government (MSIT) (No. 2020R1A2C2010200, No. 2021R1A4A1033486).
\end{acks}

\bibliographystyle{ACM-Reference-Format}
\bibliography{sample-base}


\begin{thebibliography}{38}


\ifx \showCODEN    \undefined \def \showCODEN     #1{\unskip}     \fi
\ifx \showDOI      \undefined \def \showDOI       #1{#1}\fi
\ifx \showISBNx    \undefined \def \showISBNx     #1{\unskip}     \fi
\ifx \showISBNxiii \undefined \def \showISBNxiii  #1{\unskip}     \fi
\ifx \showISSN     \undefined \def \showISSN      #1{\unskip}     \fi
\ifx \showLCCN     \undefined \def \showLCCN      #1{\unskip}     \fi
\ifx \shownote     \undefined \def \shownote      #1{#1}          \fi
\ifx \showarticletitle \undefined \def \showarticletitle #1{#1}   \fi
\ifx \showURL      \undefined \def \showURL       {\relax}        \fi
\providecommand\bibfield[2]{#2}
\providecommand\bibinfo[2]{#2}
\providecommand\natexlab[1]{#1}
\providecommand\showeprint[2][]{arXiv:#2}

\bibitem[Cen et~al\mbox{.}(2019)]%
        {cen2019representation}
\bibfield{author}{\bibinfo{person}{Yukuo Cen}, \bibinfo{person}{Xu Zou},
  \bibinfo{person}{Jianwei Zhang}, \bibinfo{person}{Hongxia Yang},
  \bibinfo{person}{Jingren Zhou}, {and} \bibinfo{person}{Jie Tang}.}
  \bibinfo{year}{2019}\natexlab{}.
\newblock \showarticletitle{Representation learning for attributed multiplex
  heterogeneous network}. In \bibinfo{booktitle}{\emph{Proceedings of the 25th
  ACM SIGKDD International Conference on Knowledge Discovery \& Data Mining}}.
  \bibinfo{pages}{1358--1368}.
\newblock


\bibitem[Defferrard et~al\mbox{.}(2016)]%
        {defferrard2016convolutional}
\bibfield{author}{\bibinfo{person}{Micha{\"e}l Defferrard},
  \bibinfo{person}{Xavier Bresson}, {and} \bibinfo{person}{Pierre
  Vandergheynst}.} \bibinfo{year}{2016}\natexlab{}.
\newblock \showarticletitle{Convolutional neural networks on graphs with fast
  localized spectral filtering}.
\newblock \bibinfo{journal}{\emph{Advances in neural information processing
  systems}}  \bibinfo{volume}{29} (\bibinfo{year}{2016}).
\newblock


\bibitem[Dong et~al\mbox{.}(2017)]%
        {dong2017metapath2vec}
\bibfield{author}{\bibinfo{person}{Yuxiao Dong}, \bibinfo{person}{Nitesh~V
  Chawla}, {and} \bibinfo{person}{Ananthram Swami}.}
  \bibinfo{year}{2017}\natexlab{}.
\newblock \showarticletitle{metapath2vec: Scalable representation learning for
  heterogeneous networks}. In \bibinfo{booktitle}{\emph{Proceedings of the 23rd
  ACM SIGKDD international conference on knowledge discovery and data mining}}.
  \bibinfo{pages}{135--144}.
\newblock


\bibitem[Estevez et~al\mbox{.}(2009)]%
        {4749258}
\bibfield{author}{\bibinfo{person}{Pablo~A. Estevez}, \bibinfo{person}{Michel
  Tesmer}, \bibinfo{person}{Claudio~A. Perez}, {and} \bibinfo{person}{Jacek~M.
  Zurada}.} \bibinfo{year}{2009}\natexlab{}.
\newblock \showarticletitle{Normalized Mutual Information Feature Selection}.
\newblock \bibinfo{journal}{\emph{IEEE Transactions on Neural Networks}}
  \bibinfo{volume}{20}, \bibinfo{number}{2} (\bibinfo{year}{2009}),
  \bibinfo{pages}{189--201}.
\newblock
\urldef\tempurl%
\url{https://doi.org/10.1109/TNN.2008.2005601}
\showDOI{\tempurl}


\bibitem[Fu et~al\mbox{.}(2017)]%
        {fu2017hin2vec}
\bibfield{author}{\bibinfo{person}{Tao-yang Fu}, \bibinfo{person}{Wang-Chien
  Lee}, {and} \bibinfo{person}{Zhen Lei}.} \bibinfo{year}{2017}\natexlab{}.
\newblock \showarticletitle{Hin2vec: Explore meta-paths in heterogeneous
  information networks for representation learning}. In
  \bibinfo{booktitle}{\emph{Proceedings of the 2017 ACM on Conference on
  Information and Knowledge Management}}. \bibinfo{pages}{1797--1806}.
\newblock


\bibitem[Fu et~al\mbox{.}(2020)]%
        {fu2020magnn}
\bibfield{author}{\bibinfo{person}{Xinyu Fu}, \bibinfo{person}{Jiani Zhang},
  \bibinfo{person}{Ziqiao Meng}, {and} \bibinfo{person}{Irwin King}.}
  \bibinfo{year}{2020}\natexlab{}.
\newblock \showarticletitle{Magnn: Metapath aggregated graph neural network for
  heterogeneous graph embedding}. In \bibinfo{booktitle}{\emph{Proceedings of
  The Web Conference 2020}}. \bibinfo{pages}{2331--2341}.
\newblock


\bibitem[Fu et~al\mbox{.}(2019)]%
        {fu2019efficient}
\bibfield{author}{\bibinfo{person}{Yanjie Fu}, \bibinfo{person}{Pengyang Wang},
  \bibinfo{person}{Jiadi Du}, \bibinfo{person}{Le Wu}, {and}
  \bibinfo{person}{Xiaolin Li}.} \bibinfo{year}{2019}\natexlab{}.
\newblock \showarticletitle{Efficient region embedding with multi-view spatial
  networks: A perspective of locality-constrained spatial autocorrelations}. In
  \bibinfo{booktitle}{\emph{Proceedings of the AAAI Conference on Artificial
  Intelligence}}, Vol.~\bibinfo{volume}{33}. \bibinfo{pages}{906--913}.
\newblock


\bibitem[Grover and Leskovec(2016)]%
        {grover2016node2vec}
\bibfield{author}{\bibinfo{person}{Aditya Grover} {and} \bibinfo{person}{Jure
  Leskovec}.} \bibinfo{year}{2016}\natexlab{}.
\newblock \showarticletitle{node2vec: Scalable feature learning for networks}.
  In \bibinfo{booktitle}{\emph{Proceedings of the 22nd ACM SIGKDD international
  conference on Knowledge discovery and data mining}}.
  \bibinfo{pages}{855--864}.
\newblock


\bibitem[Hamilton et~al\mbox{.}(2017)]%
        {hamilton2017inductive}
\bibfield{author}{\bibinfo{person}{Will Hamilton}, \bibinfo{person}{Zhitao
  Ying}, {and} \bibinfo{person}{Jure Leskovec}.}
  \bibinfo{year}{2017}\natexlab{}.
\newblock \showarticletitle{Inductive representation learning on large graphs}.
\newblock \bibinfo{journal}{\emph{Advances in neural information processing
  systems}}  \bibinfo{volume}{30} (\bibinfo{year}{2017}).
\newblock


\bibitem[Hubert and Arabie(1985)]%
        {Hubert1985ComparingP}
\bibfield{author}{\bibinfo{person}{Lawrence~J. Hubert} {and}
  \bibinfo{person}{Phipps Arabie}.} \bibinfo{year}{1985}\natexlab{}.
\newblock \showarticletitle{Comparing partitions}.
\newblock \bibinfo{journal}{\emph{Journal of Classification}}
  \bibinfo{volume}{2} (\bibinfo{year}{1985}), \bibinfo{pages}{193--218}.
\newblock


\bibitem[Hui et~al\mbox{.}(2020)]%
        {hui2020predicting}
\bibfield{author}{\bibinfo{person}{Bo Hui}, \bibinfo{person}{Da Yan},
  \bibinfo{person}{Wei-Shinn Ku}, {and} \bibinfo{person}{Wenlu Wang}.}
  \bibinfo{year}{2020}\natexlab{}.
\newblock \showarticletitle{Predicting economic growth by region embedding: A
  multigraph convolutional network approach}. In
  \bibinfo{booktitle}{\emph{Proceedings of the 29th ACM International
  Conference on Information \& Knowledge Management}}.
  \bibinfo{pages}{555--564}.
\newblock


\bibitem[Jean et~al\mbox{.}(2019)]%
        {jean2019tile2vec}
\bibfield{author}{\bibinfo{person}{Neal Jean}, \bibinfo{person}{Sherrie Wang},
  \bibinfo{person}{Anshul Samar}, \bibinfo{person}{George Azzari},
  \bibinfo{person}{David Lobell}, {and} \bibinfo{person}{Stefano Ermon}.}
  \bibinfo{year}{2019}\natexlab{}.
\newblock \showarticletitle{Tile2vec: Unsupervised representation learning for
  spatially distributed data}. In \bibinfo{booktitle}{\emph{Proceedings of the
  AAAI Conference on Artificial Intelligence}}, Vol.~\bibinfo{volume}{33}.
  \bibinfo{pages}{3967--3974}.
\newblock


\bibitem[Kailath(1967)]%
        {1089532}
\bibfield{author}{\bibinfo{person}{T. Kailath}.}
  \bibinfo{year}{1967}\natexlab{}.
\newblock \showarticletitle{The Divergence and Bhattacharyya Distance Measures
  in Signal Selection}.
\newblock \bibinfo{journal}{\emph{IEEE Transactions on Communication
  Technology}} \bibinfo{volume}{15}, \bibinfo{number}{1}
  (\bibinfo{year}{1967}), \bibinfo{pages}{52--60}.
\newblock
\urldef\tempurl%
\url{https://doi.org/10.1109/TCOM.1967.1089532}
\showDOI{\tempurl}


\bibitem[Kipf and Welling(2016)]%
        {kipf2016semi}
\bibfield{author}{\bibinfo{person}{Thomas~N Kipf} {and} \bibinfo{person}{Max
  Welling}.} \bibinfo{year}{2016}\natexlab{}.
\newblock \showarticletitle{Semi-supervised classification with graph
  convolutional networks}.
\newblock \bibinfo{journal}{\emph{arXiv preprint arXiv:1609.02907}}
  (\bibinfo{year}{2016}).
\newblock


\bibitem[Kullback and Leibler(1951)]%
        {Kullback1951OnIA}
\bibfield{author}{\bibinfo{person}{Solomon Kullback} {and}
  \bibinfo{person}{R.~A. Leibler}.} \bibinfo{year}{1951}\natexlab{}.
\newblock \showarticletitle{On Information and Sufficiency}.
\newblock \bibinfo{journal}{\emph{Annals of Mathematical Statistics}}
  \bibinfo{volume}{22} (\bibinfo{year}{1951}), \bibinfo{pages}{79--86}.
\newblock


\bibitem[Liu et~al\mbox{.}(2020)]%
        {liu2020learning}
\bibfield{author}{\bibinfo{person}{Zhicheng Liu}, \bibinfo{person}{Fabio
  Miranda}, \bibinfo{person}{Weiting Xiong}, \bibinfo{person}{Junyan Yang},
  \bibinfo{person}{Qiao Wang}, {and} \bibinfo{person}{Claudio Silva}.}
  \bibinfo{year}{2020}\natexlab{}.
\newblock \showarticletitle{Learning geo-contextual embeddings for commuting
  flow prediction}. In \bibinfo{booktitle}{\emph{Proceedings of the AAAI
  Conference on Artificial Intelligence}}, Vol.~\bibinfo{volume}{34}.
  \bibinfo{pages}{808--816}.
\newblock


\bibitem[Luxen and Vetter(2011)]%
        {luxen-vetter-2011}
\bibfield{author}{\bibinfo{person}{Dennis Luxen} {and}
  \bibinfo{person}{Christian Vetter}.} \bibinfo{year}{2011}\natexlab{}.
\newblock \showarticletitle{Real-time routing with OpenStreetMap data}. In
  \bibinfo{booktitle}{\emph{Proceedings of the 19th ACM SIGSPATIAL
  International Conference on Advances in Geographic Information Systems}}
  (Chicago, Illinois) \emph{(\bibinfo{series}{GIS '11})}.
  \bibinfo{publisher}{ACM}, \bibinfo{address}{New York, NY, USA},
  \bibinfo{pages}{513--516}.
\newblock
\showISBNx{978-1-4503-1031-4}
\urldef\tempurl%
\url{https://doi.org/10.1145/2093973.2094062}
\showDOI{\tempurl}


\bibitem[Macqueen(1967)]%
        {Macqueen67somemethods}
\bibfield{author}{\bibinfo{person}{J. Macqueen}.}
  \bibinfo{year}{1967}\natexlab{}.
\newblock \showarticletitle{Some methods for classification and analysis of
  multivariate observations}. In \bibinfo{booktitle}{\emph{In 5-th Berkeley
  Symposium on Mathematical Statistics and Probability}}.
  \bibinfo{pages}{281--297}.
\newblock


\bibitem[Perozzi et~al\mbox{.}(2014)]%
        {perozzi2014deepwalk}
\bibfield{author}{\bibinfo{person}{Bryan Perozzi}, \bibinfo{person}{Rami
  Al-Rfou}, {and} \bibinfo{person}{Steven Skiena}.}
  \bibinfo{year}{2014}\natexlab{}.
\newblock \showarticletitle{Deepwalk: Online learning of social
  representations}. In \bibinfo{booktitle}{\emph{Proceedings of the 20th ACM
  SIGKDD international conference on Knowledge discovery and data mining}}.
  \bibinfo{pages}{701--710}.
\newblock


\bibitem[Qu et~al\mbox{.}(2017)]%
        {qu2017attention}
\bibfield{author}{\bibinfo{person}{Meng Qu}, \bibinfo{person}{Jian Tang},
  \bibinfo{person}{Jingbo Shang}, \bibinfo{person}{Xiang Ren},
  \bibinfo{person}{Ming Zhang}, {and} \bibinfo{person}{Jiawei Han}.}
  \bibinfo{year}{2017}\natexlab{}.
\newblock \showarticletitle{An attention-based collaboration framework for
  multi-view network representation learning}. In
  \bibinfo{booktitle}{\emph{Proceedings of the 2017 ACM on Conference on
  Information and Knowledge Management}}. \bibinfo{pages}{1767--1776}.
\newblock


\bibitem[Shang et~al\mbox{.}(2016)]%
        {shang2016meta}
\bibfield{author}{\bibinfo{person}{Jingbo Shang}, \bibinfo{person}{Meng Qu},
  \bibinfo{person}{Jialu Liu}, \bibinfo{person}{Lance~M Kaplan},
  \bibinfo{person}{Jiawei Han}, {and} \bibinfo{person}{Jian Peng}.}
  \bibinfo{year}{2016}\natexlab{}.
\newblock \showarticletitle{Meta-path guided embedding for similarity search in
  large-scale heterogeneous information networks}.
\newblock \bibinfo{journal}{\emph{arXiv preprint arXiv:1610.09769}}
  (\bibinfo{year}{2016}).
\newblock


\bibitem[Shi et~al\mbox{.}(2018a)]%
        {shi2018heterogeneous}
\bibfield{author}{\bibinfo{person}{Chuan Shi}, \bibinfo{person}{Binbin Hu},
  \bibinfo{person}{Wayne~Xin Zhao}, {and} \bibinfo{person}{S~Yu Philip}.}
  \bibinfo{year}{2018}\natexlab{a}.
\newblock \showarticletitle{Heterogeneous information network embedding for
  recommendation}.
\newblock \bibinfo{journal}{\emph{IEEE Transactions on Knowledge and Data
  Engineering}} \bibinfo{volume}{31}, \bibinfo{number}{2}
  (\bibinfo{year}{2018}), \bibinfo{pages}{357--370}.
\newblock


\bibitem[Shi et~al\mbox{.}(2018b)]%
        {shi2018easing}
\bibfield{author}{\bibinfo{person}{Yu Shi}, \bibinfo{person}{Qi Zhu},
  \bibinfo{person}{Fang Guo}, \bibinfo{person}{Chao Zhang}, {and}
  \bibinfo{person}{Jiawei Han}.} \bibinfo{year}{2018}\natexlab{b}.
\newblock \showarticletitle{Easing embedding learning by comprehensive
  transcription of heterogeneous information networks}. In
  \bibinfo{booktitle}{\emph{Proceedings of the 24th ACM SIGKDD International
  Conference on Knowledge Discovery \& Data Mining}}.
  \bibinfo{pages}{2190--2199}.
\newblock


\bibitem[Sun and Han(2013)]%
        {sun2013mining}
\bibfield{author}{\bibinfo{person}{Yizhou Sun} {and} \bibinfo{person}{Jiawei
  Han}.} \bibinfo{year}{2013}\natexlab{}.
\newblock \showarticletitle{Mining heterogeneous information networks: a
  structural analysis approach}.
\newblock \bibinfo{journal}{\emph{Acm Sigkdd Explorations Newsletter}}
  \bibinfo{volume}{14}, \bibinfo{number}{2} (\bibinfo{year}{2013}),
  \bibinfo{pages}{20--28}.
\newblock


\bibitem[Sun et~al\mbox{.}(2011)]%
        {sun2011pathsim}
\bibfield{author}{\bibinfo{person}{Yizhou Sun}, \bibinfo{person}{Jiawei Han},
  \bibinfo{person}{Xifeng Yan}, \bibinfo{person}{Philip~S Yu}, {and}
  \bibinfo{person}{Tianyi Wu}.} \bibinfo{year}{2011}\natexlab{}.
\newblock \showarticletitle{Pathsim: Meta path-based top-k similarity search in
  heterogeneous information networks}.
\newblock \bibinfo{journal}{\emph{Proceedings of the VLDB Endowment}}
  \bibinfo{volume}{4}, \bibinfo{number}{11} (\bibinfo{year}{2011}),
  \bibinfo{pages}{992--1003}.
\newblock


\bibitem[Tang et~al\mbox{.}(2015)]%
        {tang2015line}
\bibfield{author}{\bibinfo{person}{Jian Tang}, \bibinfo{person}{Meng Qu},
  \bibinfo{person}{Mingzhe Wang}, \bibinfo{person}{Ming Zhang},
  \bibinfo{person}{Jun Yan}, {and} \bibinfo{person}{Qiaozhu Mei}.}
  \bibinfo{year}{2015}\natexlab{}.
\newblock \showarticletitle{Line: Large-scale information network embedding}.
  In \bibinfo{booktitle}{\emph{Proceedings of the 24th international conference
  on world wide web}}. \bibinfo{pages}{1067--1077}.
\newblock


\bibitem[Veli{\v{c}}kovi{\'c} et~al\mbox{.}(2017)]%
        {velivckovic2017graph}
\bibfield{author}{\bibinfo{person}{Petar Veli{\v{c}}kovi{\'c}},
  \bibinfo{person}{Guillem Cucurull}, \bibinfo{person}{Arantxa Casanova},
  \bibinfo{person}{Adriana Romero}, \bibinfo{person}{Pietro Lio}, {and}
  \bibinfo{person}{Yoshua Bengio}.} \bibinfo{year}{2017}\natexlab{}.
\newblock \showarticletitle{Graph attention networks}.
\newblock \bibinfo{journal}{\emph{arXiv preprint arXiv:1710.10903}}
  (\bibinfo{year}{2017}).
\newblock


\bibitem[Wang and Li(2017)]%
        {wang2017region}
\bibfield{author}{\bibinfo{person}{Hongjian Wang} {and}
  \bibinfo{person}{Zhenhui Li}.} \bibinfo{year}{2017}\natexlab{}.
\newblock \showarticletitle{Region representation learning via mobility flow}.
  In \bibinfo{booktitle}{\emph{Proceedings of the 2017 ACM on Conference on
  Information and Knowledge Management}}. \bibinfo{pages}{237--246}.
\newblock


\bibitem[Wang et~al\mbox{.}(2019b)]%
        {wang2019dgl}
\bibfield{author}{\bibinfo{person}{Minjie Wang}, \bibinfo{person}{Da Zheng},
  \bibinfo{person}{Zihao Ye}, \bibinfo{person}{Quan Gan},
  \bibinfo{person}{Mufei Li}, \bibinfo{person}{Xiang Song},
  \bibinfo{person}{Jinjing Zhou}, \bibinfo{person}{Chao Ma},
  \bibinfo{person}{Lingfan Yu}, \bibinfo{person}{Yu Gai},
  \bibinfo{person}{Tianjun Xiao}, \bibinfo{person}{Tong He},
  \bibinfo{person}{George Karypis}, \bibinfo{person}{Jinyang Li}, {and}
  \bibinfo{person}{Zheng Zhang}.} \bibinfo{year}{2019}\natexlab{b}.
\newblock \showarticletitle{Deep Graph Library: A Graph-Centric,
  Highly-Performant Package for Graph Neural Networks}.
\newblock \bibinfo{journal}{\emph{arXiv preprint arXiv:1909.01315}}
  (\bibinfo{year}{2019}).
\newblock


\bibitem[Wang et~al\mbox{.}(2019a)]%
        {wang2019heterogeneous}
\bibfield{author}{\bibinfo{person}{Xiao Wang}, \bibinfo{person}{Houye Ji},
  \bibinfo{person}{Chuan Shi}, \bibinfo{person}{Bai Wang},
  \bibinfo{person}{Yanfang Ye}, \bibinfo{person}{Peng Cui}, {and}
  \bibinfo{person}{Philip~S Yu}.} \bibinfo{year}{2019}\natexlab{a}.
\newblock \showarticletitle{Heterogeneous graph attention network}. In
  \bibinfo{booktitle}{\emph{The world wide web conference}}.
  \bibinfo{pages}{2022--2032}.
\newblock


\bibitem[Wang et~al\mbox{.}(2020)]%
        {wang2020urban2vec}
\bibfield{author}{\bibinfo{person}{Zhecheng Wang}, \bibinfo{person}{Haoyuan
  Li}, {and} \bibinfo{person}{Ram Rajagopal}.} \bibinfo{year}{2020}\natexlab{}.
\newblock \showarticletitle{Urban2vec: Incorporating street view imagery and
  pois for multi-modal urban neighborhood embedding}. In
  \bibinfo{booktitle}{\emph{Proceedings of the AAAI Conference on Artificial
  Intelligence}}, Vol.~\bibinfo{volume}{34}. \bibinfo{pages}{1013--1020}.
\newblock


\bibitem[Xu et~al\mbox{.}(2018)]%
        {xu2018powerful}
\bibfield{author}{\bibinfo{person}{Keyulu Xu}, \bibinfo{person}{Weihua Hu},
  \bibinfo{person}{Jure Leskovec}, {and} \bibinfo{person}{Stefanie Jegelka}.}
  \bibinfo{year}{2018}\natexlab{}.
\newblock \showarticletitle{How powerful are graph neural networks?}
\newblock \bibinfo{journal}{\emph{arXiv preprint arXiv:1810.00826}}
  (\bibinfo{year}{2018}).
\newblock


\bibitem[Yang et~al\mbox{.}(2016)]%
        {10.1145/2814575}
\bibfield{author}{\bibinfo{person}{Dingqi Yang}, \bibinfo{person}{Daqing
  Zhang}, {and} \bibinfo{person}{Bingqing Qu}.}
  \bibinfo{year}{2016}\natexlab{}.
\newblock \showarticletitle{Participatory Cultural Mapping Based on Collective
  Behavior Data in Location-Based Social Networks}.
\newblock  \bibinfo{volume}{7}, \bibinfo{number}{3} (\bibinfo{year}{2016}).
\newblock
\showISSN{2157-6904}
\urldef\tempurl%
\url{https://doi.org/10.1145/2814575}
\showDOI{\tempurl}


\bibitem[Yao et~al\mbox{.}(2018)]%
        {yao2018representing}
\bibfield{author}{\bibinfo{person}{Zijun Yao}, \bibinfo{person}{Yanjie Fu},
  \bibinfo{person}{Bin Liu}, \bibinfo{person}{Wangsu Hu}, {and}
  \bibinfo{person}{Hui Xiong}.} \bibinfo{year}{2018}\natexlab{}.
\newblock \showarticletitle{Representing urban functions through zone embedding
  with human mobility patterns}. In \bibinfo{booktitle}{\emph{Proceedings of
  the Twenty-Seventh International Joint Conference on Artificial Intelligence
  (IJCAI-18)}}.
\newblock


\bibitem[Zhang et~al\mbox{.}(2017)]%
        {zhang2017regions}
\bibfield{author}{\bibinfo{person}{Chao Zhang}, \bibinfo{person}{Keyang Zhang},
  \bibinfo{person}{Quan Yuan}, \bibinfo{person}{Haoruo Peng},
  \bibinfo{person}{Yu Zheng}, \bibinfo{person}{Tim Hanratty},
  \bibinfo{person}{Shaowen Wang}, {and} \bibinfo{person}{Jiawei Han}.}
  \bibinfo{year}{2017}\natexlab{}.
\newblock \showarticletitle{Regions, periods, activities: Uncovering urban
  dynamics via cross-modal representation learning}. In
  \bibinfo{booktitle}{\emph{Proceedings of the 26th International Conference on
  World Wide Web}}. \bibinfo{pages}{361--370}.
\newblock


\bibitem[Zhang et~al\mbox{.}(2018)]%
        {zhang2018scalable}
\bibfield{author}{\bibinfo{person}{Hongming Zhang}, \bibinfo{person}{Liwei
  Qiu}, \bibinfo{person}{Lingling Yi}, {and} \bibinfo{person}{Yangqiu Song}.}
  \bibinfo{year}{2018}\natexlab{}.
\newblock \showarticletitle{Scalable multiplex network embedding.}. In
  \bibinfo{booktitle}{\emph{IJCAI}}, Vol.~\bibinfo{volume}{18}.
  \bibinfo{pages}{3082--3088}.
\newblock


\bibitem[Zhang et~al\mbox{.}(2021)]%
        {zhang2021multi}
\bibfield{author}{\bibinfo{person}{Mingyang Zhang}, \bibinfo{person}{Tong Li},
  \bibinfo{person}{Yong Li}, {and} \bibinfo{person}{Pan Hui}.}
  \bibinfo{year}{2021}\natexlab{}.
\newblock \showarticletitle{Multi-view joint graph representation learning for
  urban region embedding}. In \bibinfo{booktitle}{\emph{Proceedings of the
  Twenty-Ninth International Conference on International Joint Conferences on
  Artificial Intelligence}}. \bibinfo{pages}{4431--4437}.
\newblock


\bibitem[Zhang et~al\mbox{.}(2019)]%
        {zhang2019unifying}
\bibfield{author}{\bibinfo{person}{Yunchao Zhang}, \bibinfo{person}{Yanjie Fu},
  \bibinfo{person}{Pengyang Wang}, \bibinfo{person}{Xiaolin Li}, {and}
  \bibinfo{person}{Yu Zheng}.} \bibinfo{year}{2019}\natexlab{}.
\newblock \showarticletitle{Unifying inter-region autocorrelation and
  intra-region structures for spatial embedding via collective adversarial
  learning}. In \bibinfo{booktitle}{\emph{Proceedings of the 25th ACM SIGKDD
  International Conference on Knowledge Discovery \& Data Mining}}.
  \bibinfo{pages}{1700--1708}.
\newblock


\end{thebibliography}

\end{document}